\documentclass{article}

\usepackage[dblblindworkshop, final]{neurips_2026}

\usepackage[utf8]{inputenc} 
\usepackage[T1]{fontenc}    
\usepackage{hyperref}       
\usepackage{url}            
\usepackage{booktabs}       
\usepackage{amsfonts}       
\usepackage{nicefrac}       
\usepackage{microtype}      
\usepackage{xcolor}         
\usepackage{graphicx}
\usepackage{svg}
\usepackage{amsmath}
\usepackage{todonotes}
\usepackage{natbib}
\usepackage{enumitem}
\setlist{nosep, topsep=2pt}
\title{The role of injected noise in diffusion model generalization}
\workshoptitle{2nd Workshop on Principles of Generative Modeling (PriGM)}

\author{
  Vincent Emmerling$^*$, Felix G\"unther$^*$ and David Kappel \\
  CITEC, Bielefeld University\\
  Bielefeld, Germany \\
  \texttt{\{vincent.emmerling, felix.guenther, david.kappel\}@uni-bielefeld.de} \\ [1mm]
  $^{*}$Equal contribution
}

\begin{document}

\maketitle

\begin{abstract}
  Diffusion models trained on distinct datasets can produce highly similar samples when prompted with the same noise input, suggesting that they learn a generalized data distribution. We investigate the role of noise injected during the sampling process, and show that it plays an important role in making or breaking this strong generalization effect. Based on this finding we investigate the robustness of sample similarity to changes in the injected noise, and find that its impact is largest in the sampling regime where broad structure already exists but the precise identity of the sample is not yet determined.
\end{abstract}
\begin{figure}[htbp]
    \centering
    \includegraphics[width=\linewidth]{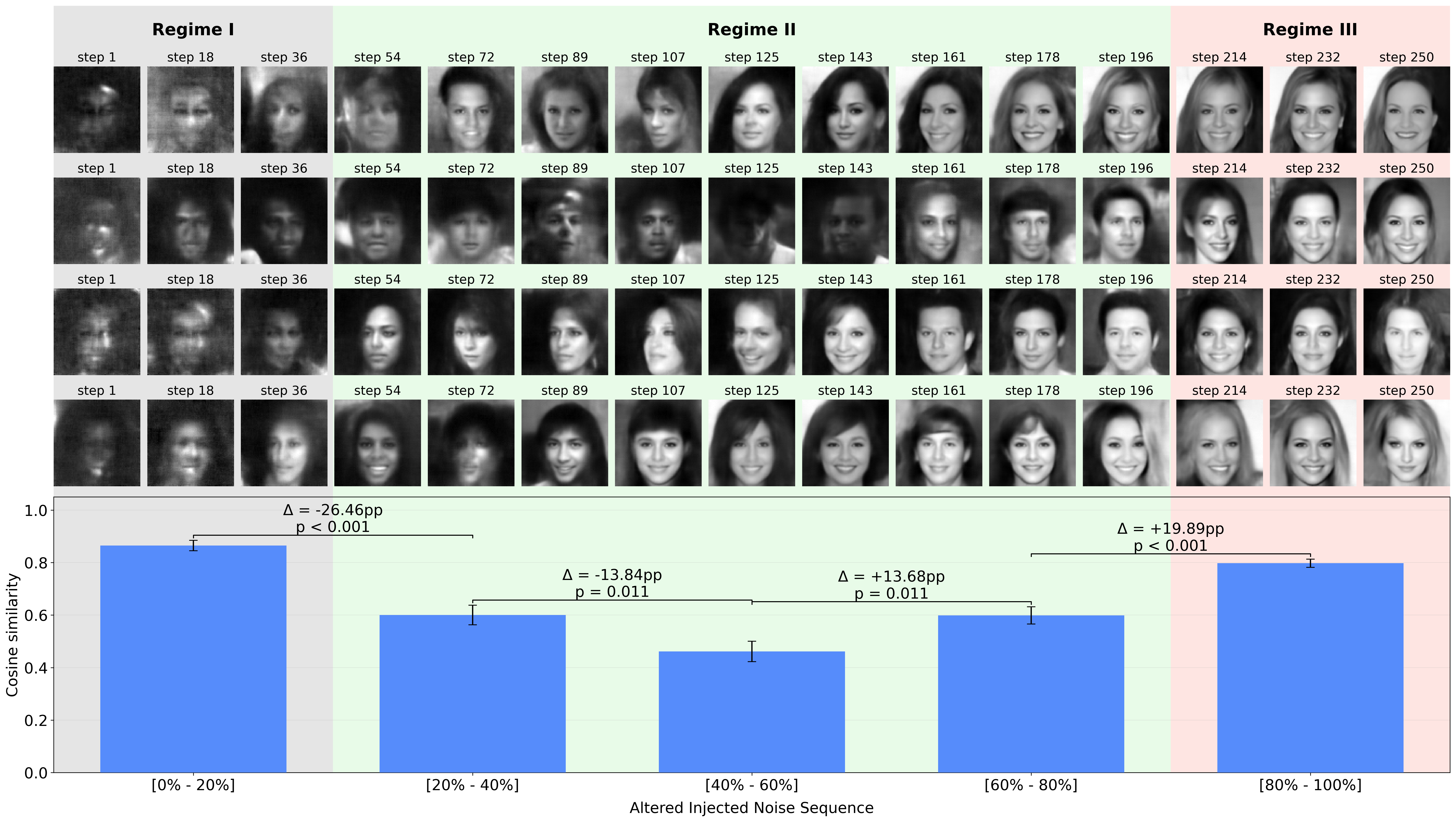}
    \caption{
    \textbf{Top}: Exemplary denoised sampling trajectories (rows) using different injected noise. \textbf{Bottom}: The sensitivity to changes in the injected noise sequence is significantly higher during regime II than regime I and III ($p$-values show two-sided paired $t$-tests for adjacent intervals, $\alpha = 0.05$).}
    \label{fig:experiment_3_regimes_trajectories}
\end{figure}

\section{Introduction}
In recent years, generative models have made significant progress, producing results of such high fidelity that they are increasingly difficult to distinguish from the training samples \citep{harshvardhan2020comprehensive, cao2024survey}. The development of diffusion models (DMs) has been instrumental in improving the quality of modern generative models, particularly in domains such as image and video generation \citep{sohl2015deep, song2019generative, saharia2022photorealistic, bar2024lumiere}. This development has also sparked interest in understanding the mechanisms that enable generalization and consistency in DMs. \cite{kadkhodaie2024generalization} showed that the deep neural networks (DNNs) on which DMs are based are capable of strong generalization. When presented with the same noise vector, two DMs trained on different subsets of training data produce highly similar images. The study found that inductive biases within DNNs are the main regulariser that enables this strong generalization. However, as they used the same injected noise during the diffusion sampling process — i.e. the same starting noise sample \emph{and} the entire injected noise sequence — it remains unclear exactly which part of the denoising process contributes to strong generalization. Our main findings are summarized in Figure~\ref{fig:experiment_3_regimes_trajectories}. We demonstrate that the injected noise during the central part of the sampling process has a significant impact on consistency, whereas the initial and late parts do not. This finding is consistent with recent accounts of DM dynamics showing that the speciation regime (regime II), but not regimes I or III, exhibit generalization capabilities in DMs \citep{biroli2024dynamical}. 

\section{Background}
\subsection{Diffusion sampling algorithms}
DMs can be modeled as a neural network parameterizing a vector field $u_t$ along with a diffusion coefficient $\sigma_t$. The sampling process then simulates the Stochastic Differential Equation (SDE) \citep{flowsanddiffusions2026}, for $t \in [0,T]$:
\begin{align*}
    \mathbf{x}_0 &\sim \mathcal{N}(\mathbf{0}, \mathbf{1})\,,\quad d\mathbf{x}_t = u_t(\mathbf{x}_t) + \sigma_t dW_t\,
\end{align*}
with $W$ being a Wiener process. This SDE transforms an initial Gaussian distribution into a target distribution induced by a training dataset. Many DM sampling algorithms follow this model (using a discrete time approximation) with a core update function based on the Euler-Maruyama method \citep{flowsanddiffusions2026}. Here we adapt the sampling method introduced in \citep{kadkhodaie2020solving}. We initialize $\mathbf{x}_0 \sim \mathcal{N}(\mathbf{0}, \mathbf{1})$, called the \emph{starting noise} below. Note that we follow \cite{flowsanddiffusions2026} in defining $\mathbf{x}_0$ as the starting noise, not a data sample. The update rule is then given by
\begin{align}
    \mathbf{x}_{t+1} = \mathbf{x}_t + h_tu_t(\mathbf{x}_t) + \sigma_t \mathbf{z}_t\,, \quad \mathbf{z}_t \sim \mathcal{N}(\mathbf{0}, \mathbf{1})\,, 
\end{align}
where $h_t$ determines the step size in the direction predicted by the model and $\sigma_t \in \mathbb{R}$ is the diffusion coefficient as per \cite{kadkhodaie2020solving} (See Fig.~\ref{fig:noise_schedule} in the Appendix for a typical noise schedule). We call the sequence of independent noise samples injected at each step the \emph{injected noise sequence} $(\mathbf{z}_t)_{0\leq t \leq T - 1}$ and the intermediate states computed during $T$ denoising steps the \emph{sampling trajectory} $(\mathbf{x}_t)_{0\leq t \leq T}$. We additionally define the \emph{denoised state} $\mathbf{\hat{x}}_t = \mathbf{x}_t + u_t(\mathbf{x}_t)$ as the sampling state obtained by performing a denoising step without injecting noise, i.e., with $\sigma_t = 0$ and $h_t = 1$. We use this state to inspect the direction in which the model guides the sampling trajectory.

\subsection{Related work}
\cite{kadkhodaie2024generalization} identified strong generalization in DNNs that learn nearly the
same score function even when trained on non-overlapping datasets. This is true even across different model architectures and sampling strategies \citep{zhang2024emergence}. \cite{li2024understanding} and \cite{wang2026rmtconsistency} found that shared Gaussian statistics of the data splits are an important factor in this effect.
\cite{biroli2024dynamical} found that the denoising process in DMs is characterized by three distinct regimes. Brownian motion (regime I) is followed by speciation, where models truly generalize (regime II) and finally collapse (regime III).
Memorization is avoided, since the approximate score function, learned by DNNs is smoother than the exact one \citep{biroli2024dynamical, farghly2026diffusion}. Furthermore, an implicit dynamical regularization in the training dynamics of DMs avoids memorization even in highly overparameterized settings \citep{bonnaire2026diffusion}. \cite{sahraee2026geometry} found that the noise level of $\mathbf{x}_t$ already encodes $t$, meaning that even when this information is not provided explicitly, DMs are able to infer the phase of the diffusion process.

\section{Experiments}
\label{sec:experiments}
Our experimental setup is based on the experiments by \cite{kadkhodaie2024generalization}.
For all experiments, we explicitly generate the sequence $\mathbf{z}_t$ of injected noise and pass it to the solver. Instead of a dynamic Signal-to-Noise-Ratio stopping criterion, we use a fixed $T=250$ to be able to compare the sampling trajectories more easily. We performed all experiments using the pretrained models provided by \cite{kadkhodaie2024generalization} to maximize reproducibility, i.e. we used a UNet network with 7.6M parameters trained on $10^5$ downsampled images from the CelebA dataset \citep{liu2015faceattributes}. Additional experiments on our own models are provided in Appendix~\ref{sec:butterflies}. The metric for assessing distance between generated samples is the cosine similarity, following \cite{kadkhodaie2024generalization}. All experiments were repeated for 100 independent trials to ensure statistical significance.

\subsection{Importance of the injected noise sequence}
\label{sec:injected_vs_different_shared_noise}
\begin{figure}[htbp]
    \centering
    \includegraphics[width=\linewidth]{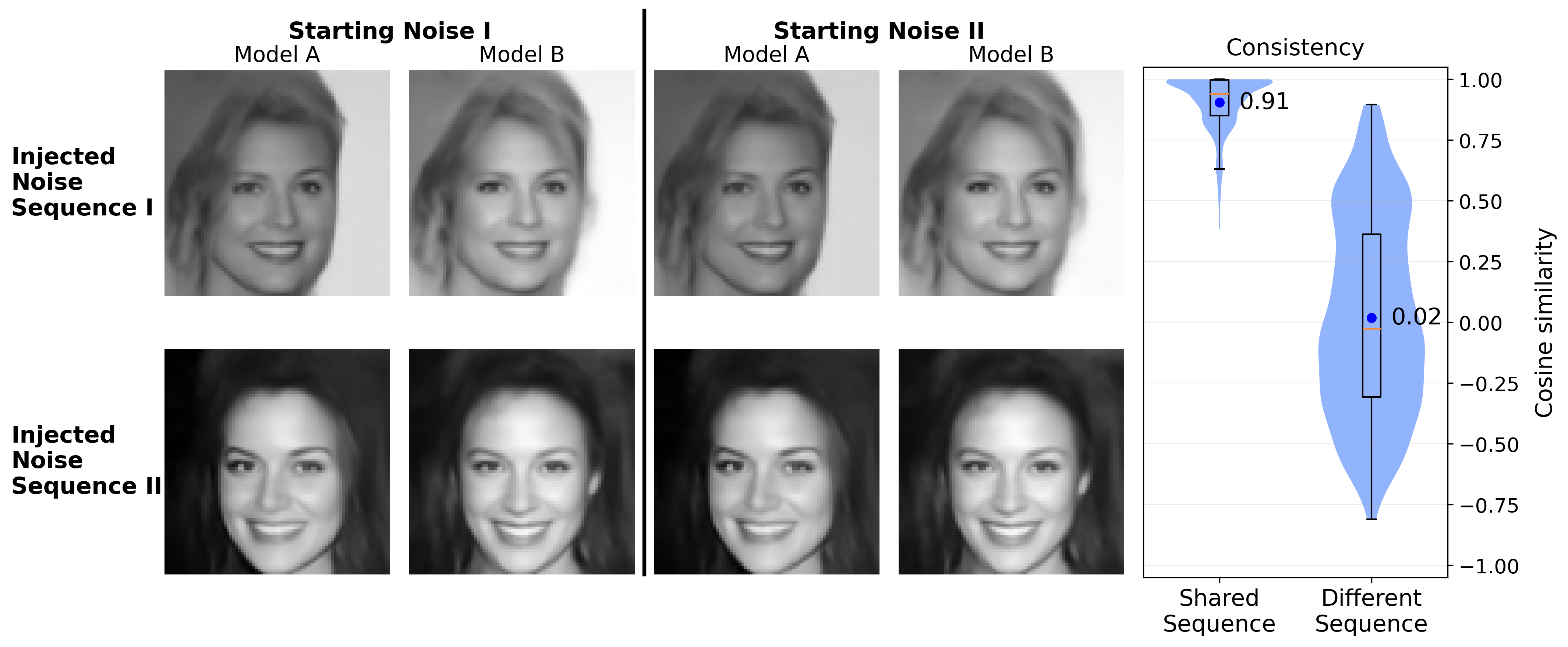}
    \caption{\textbf{Left}: Evaluating the impact of pretrained model, starting noise and injected noise sequence. Generated images are reliably robust against changes in starting noise and which pretrained model is used for sampling. The injected noise sequence has the largest impact in determining the final image. Exemplary images for one trial are shown. \textbf{Right}: To quantitatively asses consistency we report the average cosine similarity for configurations sharing/not-sharing the injected noise sequence. Configurations sharing the injected noise sequence produce highly consistent images, whereas using different noise sequences leads to substantially different results.}
    \label{fig:experiment_1}
\end{figure}

In the first experiment we aimed to quantify the impact of starting noise compared to injected noise. In the literature, the input to DDPM sampling typically consists solely of the starting noise. However, as introduced above, we also consider the injected noise sequence as a second input variable to the diffusion process. In the results (see Fig.~\ref{fig:experiment_1}) average cosine similarity is $0.91$ when using the same injected noise and $0.02$ otherwise, whereas it is 0.93 and 0.89 when using the same and different starting noise but same injected noise, respectively. This demonstrates that as long as the injected noise sequence is shared results will reliably be consistent with each other even across different models and starting noise.  

\subsection{Investigating consistency during sampling}
\label{sec:similarity_during_sampling}
\begin{figure}[htbp]
    \centering
    \includegraphics[width=\linewidth]{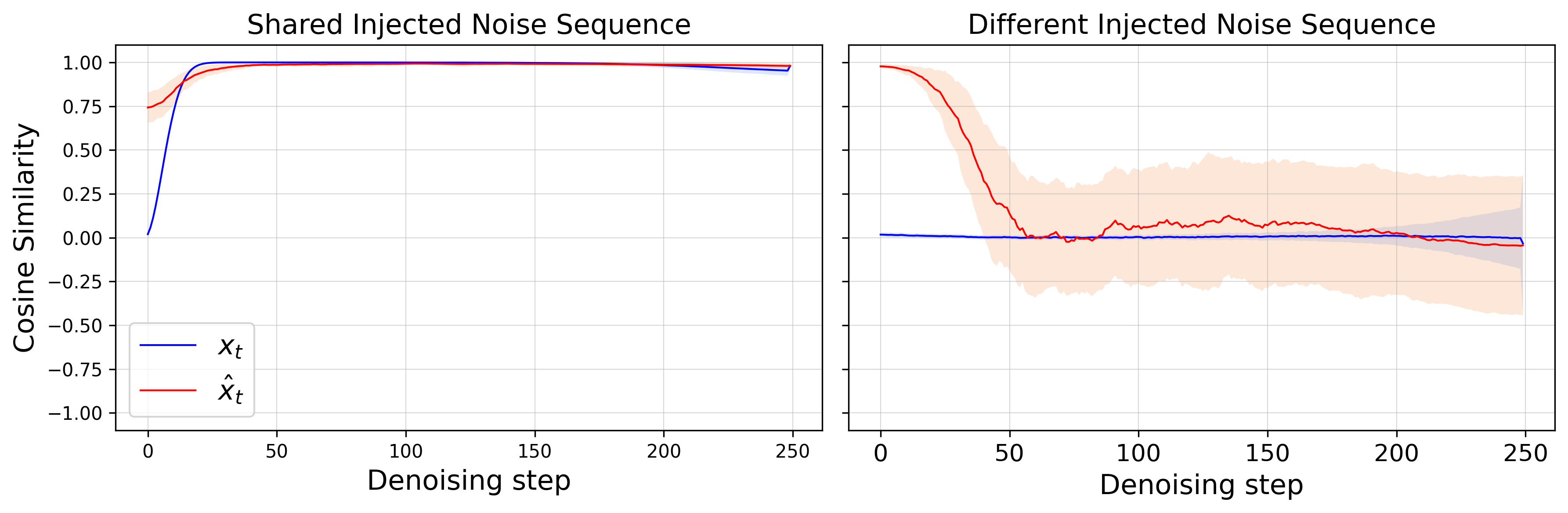}
    \caption{\textbf{Left}: Cosine similarity between intermediate states quickly rose in the beginning of the sampling process. After 10\% denoising steps it reached $\approx1.0$, staying high until sampling finished. The model could already determine the direction early on, indicated by high similarity in the beginning for $\mathbf{\hat{x}}_t$. \textbf{Right}: When the injected noise sequence was not shared the consistency stayed at $\approx0.0$ during sampling. However for $\mathbf{\hat{x}}_t$ similarity was initially high and then quickly dropped as sampling continued. We attribute this to the model producing general coarse face features early on.}
    \label{fig:experiment_cosine_similarity_during_sampling}
\end{figure}

The previous experiments demonstrated that the injected noise sequence is key to achieving consistency. Next, we investigated where along the sampling process this consistency emerges.
To do so, we generated two different sampling trajectories, using different starting noise and models, and compared the similarity of their sampling trajectories $\mathbf{x}_t$ and denoised states $\mathbf{\hat{x}}_t$. We performed this once with a shared injected noise sequence and once with different injected noise sequences. When we used the same injected noise sequence, the cosine similarity between intermediate states quickly increased to nearly $1.0$ in around 20 denoising steps (see Fig.~\ref{fig:experiment_cosine_similarity_during_sampling} Left). The picture changes dramatically where the sequence was not shared. In this case, the intermediate states did not produce consistent results, as was observed in the first experiment (see Fig.~\ref{fig:experiment_cosine_similarity_during_sampling} Right). However, an interesting observation is that even when not injecting the same noise the models still pulled states in similar directions, with $\mathbf{\hat{x}}_t$ showing high similarity during the initial 50 steps, consistent with regime I (see Fig.~\ref{fig:experiment_3_regimes_trajectories}). This experiment demonstrates that consistency is already present in the early stages of the sampling process and shows that the increase in similarity is not linear.

\subsection{Relating to dynamical regimes}
\label{sec:dynamical_regimes}
Finally, we explored the different dynamical regimes \citep{biroli2024dynamical} by 
inspecting sampling trajectories of multiple samples that do not share injected noise. In early time-steps images appear more similar before crossing the speciation boundary, consisting of very coarse features akin to mean faces (see Fig.~\ref{fig:experiment_3_regimes_trajectories}). This observation exactly aligns with the results described above, where even samples with independent injected noise produced similar outputs during the first 50 steps or 20\% of the process (see Fig.~\ref{fig:experiment_cosine_similarity_during_sampling} Right). In a similar way, we can look qualitatively for the collapse phase (see Fig.~\ref{fig:experiment_3_regimes_trajectories}), where after around 200 steps samples already look remarkably similar to the final image and for most only small changes occur, aligning with the concept of regime III. 
To verify these qualitative observations, we performed another experiment investigating the robustness of the injected noise sequence to changes in different regimes. Given that in regime I, having even entirely different injected noise produced similar model outputs (see Fig.~\ref{fig:experiment_cosine_similarity_during_sampling} Right, Fig.~\ref{fig:experiment_3_regimes_trajectories}), we expected the final samples to be robust to changes in this area. For regime II we expected less robustness as this phase of the sampling process is most crucial in determining which sample is generated. Finally, regime III was expected to have an increased robustness, as the process would already have locked in on the most important features of the sample. To test this we replaced sections of the injected noise sequence with different noise in one of five sections. Note that we chose to use five equally sized sections to ensure the magnitude of the change was consistent across the 5 setups. The results show the expected behavior of increased robustness in the first 20\% (regime I) and the last 20\% (regime III) (see Fig.~\ref{fig:experiment_3_regimes_trajectories}).

\section{Conclusion}
DMs exhibit strong generalization properties, whereby different models trained on disjoint datasets can produce highly consistent samples in response to the same sampling noise \citep{kadkhodaie2024generalization, zhang2024emergence}. Here, we demonstrate that this result is even more robust by analyzing the sensitivity to noise during the different stages of the sampling process. We found that the precise injected noise sequences most strongly influenced the generated samples during regime II of the sampling process — the regime previously identified as strongly determining the generalization capabilities of DMs. In contrast, starting noise had only a minor effect on generated samples, which is consistent with the role of the initial dynamic regime identified in \cite{biroli2024dynamical} (though see \citep{zhang2024emergence} when different deterministic samplers are used). In summary, our results further illuminate the robustness and consistency of strong generalization in diffusion models.

\section*{Acknowledgments}
David Kappel was partly funded by the Ministry of Culture and Science of the State of North Rhine-Westphalia under project SAIL (NW21-059A).

\nocite{*}
\bibliography{refs}

\section{Appendix}
The appendix contains additional experiments that extend and complement the results presented in the main paper. We first ran our experiments in a setup with a different model architecture, dataset, and sampling algorithm, providing evidence that the observed behavior is not fully dependent on the original experimental setup. We then investigated the effect of changing the initialization of the sampling process, switching from the setting of pure Gaussian noise to instead considering initial states with increasing amounts of signal from a dataset image. Together, these experiments provide further insight into the generality of the observed robustness and into how the starting conditions of the diffusion process influence the final generated sample.

\subsection{Concept of robust generalization}
\begin{figure}[htbp]
    \centering
    \includegraphics[width=\linewidth]{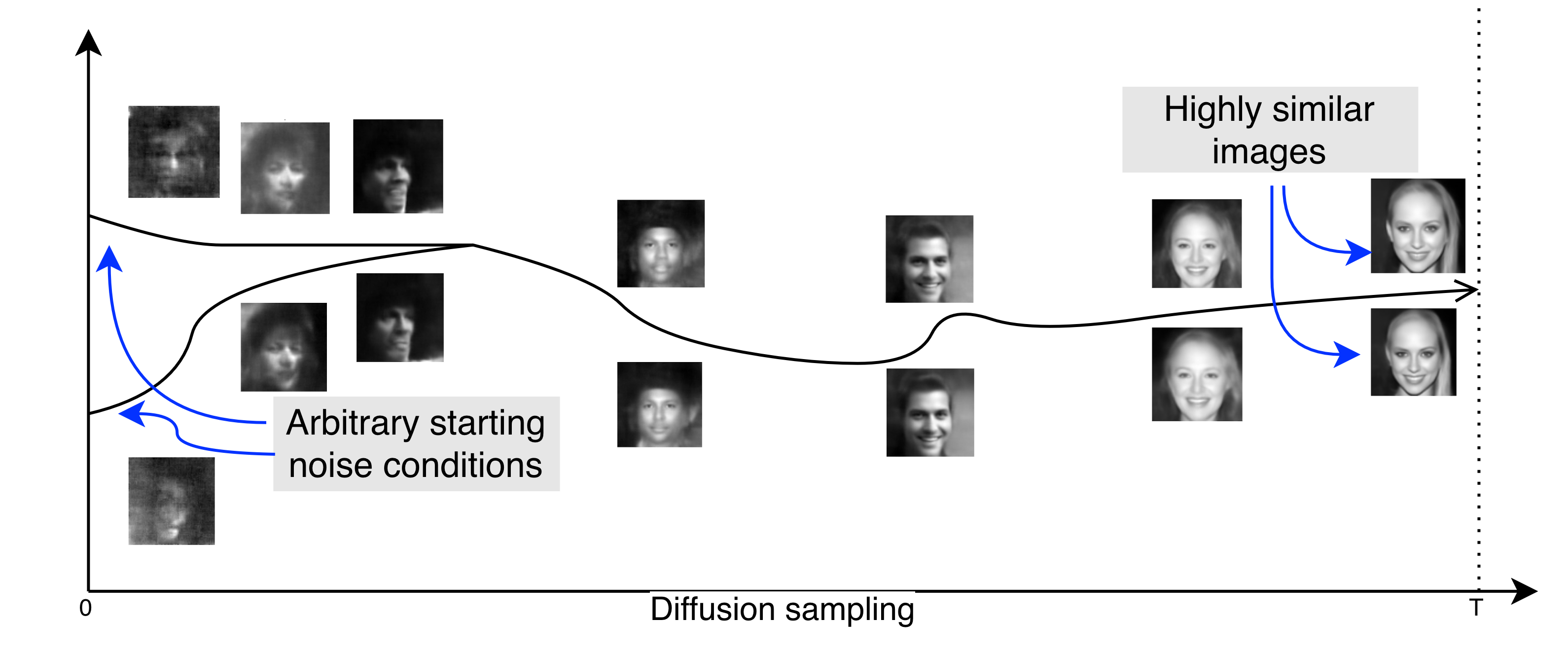}
    \caption{Generalization in diffusion models is not only dependent on starting noise. Beginning from two random starting noise conditions, intermediate states of the sampling process are shown. Similarity between samples is initially low but quickly rises. In the end both starting noise samples generate an almost identical face.}
    \label{fig:concept}
\end{figure}

\begin{figure}[htbp]
    \centering
    \includegraphics[width=\linewidth]{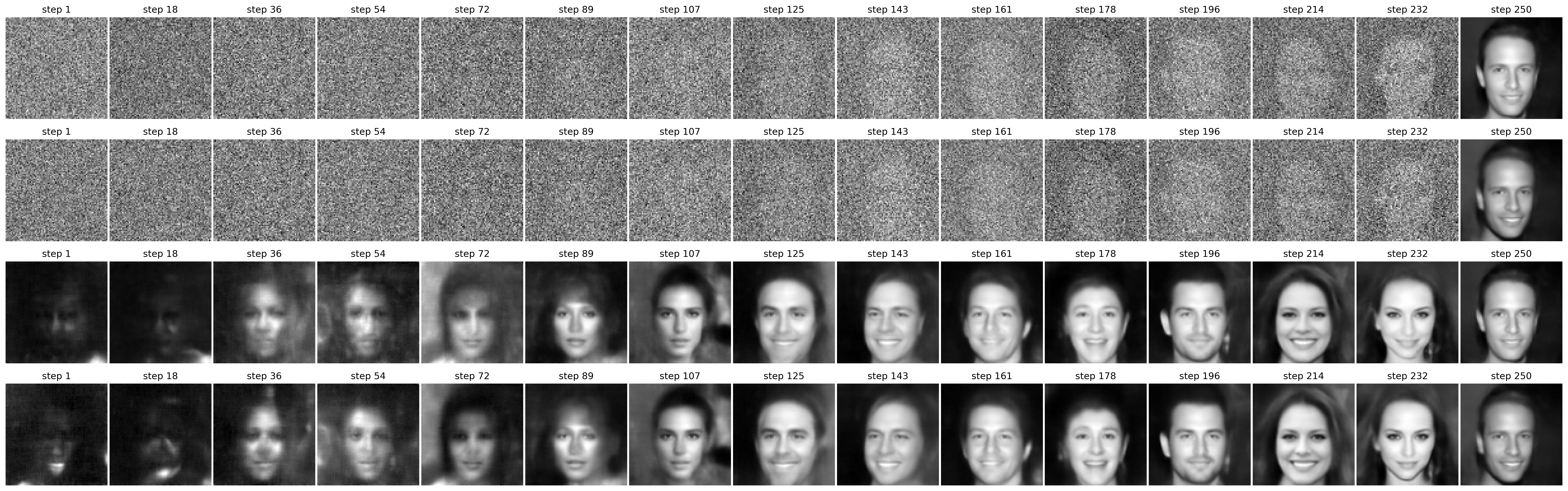}
    \caption{We show sampling trajectories for two configurations. The top two rows show intermediate states $\mathbf{x}_t$, with the first row using model A and the second row using model B, each with their own respective random starting noise. The bottom two rows show the trajectory for the same configuration and sampling process but with $\mathbf{\hat{x}}_t$, i.e. states where no injected noise is added to visually highlight the consistency between trajectory states.}
    \label{fig:trajectories_shared_injected_noise}
\end{figure}

We have shown that similarity of intermediate states $\mathbf{x}_t$ and $\mathbf{\hat{x}}_t$ quickly rises throughout sampling (see Sec.~\ref{sec:similarity_during_sampling}), here we highlight this conceptually again (see Fig.~\ref{fig:concept}). The sampling process can be initialized from arbitrary starting noise but as long as the same injected noise is used for any two initializations the final image will reliably be consistent between configurations. Notably the trajectories qualitatively show that the convergence on the final image only happens late in the process (see Sec.~\ref{sec:dynamical_regimes}) and many different faces are visible in intermediate states (see Fig.~\ref{fig:trajectories_shared_injected_noise}).

\subsection{Ablation with different model architecture and dataset}
\label{sec:butterflies}
All previous experiments used the pretrained models provided by \cite{kadkhodaie2024generalization} and the sampling procedure described in \cite{kadkhodaie2020solving}. To test if the results align across different settings, we ran our experiments again using a model trained from scratch, on a different dataset (the Smithsonian Butterflies ~\footnote{\url{https://huggingface.co/datasets/huggan/smithsonian_butterflies_subset}}) and using the standard DDPM sampling procedure \citep{ho2020denoising}. We trained two UNet models for 50 epochs each on a distinct set of 500 images. This corresponds to the setup with model A and model B, as in all experiments above (see Sec.~\ref{sec:experiments}).

For the implementation, we used the diffusers library and modified the function for noise generation to return our injected noise.

\subsubsection{Shared vs. different injected noise sequences}
\begin{figure}[htbp]
    \centering
    \includegraphics[width=\linewidth]{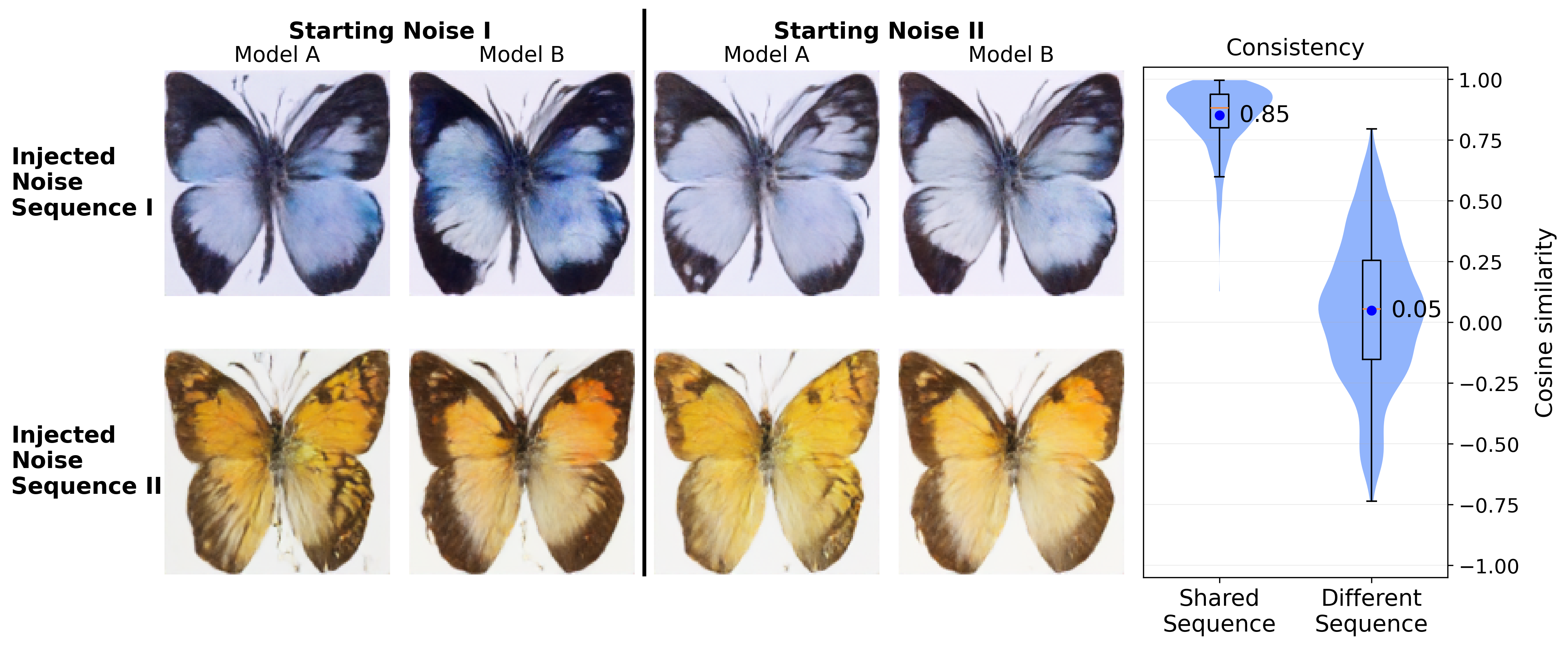}
    \caption{Generalization between model A and model B is also possible when using a much smaller dataset (500 images each) as long as the same injected noise sequence is used during DDPM sampling. This result aligns with the findings from above (see Sec.~\ref{sec:injected_vs_different_shared_noise}).}
    \label{fig:butterflies_shared_different_noise}
\end{figure}

We reran our first experiment with these models to investigate whether model A and model B can generate similar images even when the setup differs in almost all aspects. The results (see Fig.~\ref{fig:butterflies_shared_different_noise}) align with the previous results (see Fig.~\ref{fig:experiment_1}): as long as the injected noise sequence is shared, the results are reliably consistent between model A and model B. The results from \cite{kadkhodaie2024generalization} suggest that generalization can only occur if sufficiently large training sets are used. This experiment shows that even with small datasets, it is possible for models to generate highly similar images. 

\subsubsection{Rise of similarity during sampling}
\begin{figure}[htbp]
    \centering
    \includegraphics[width=\linewidth]{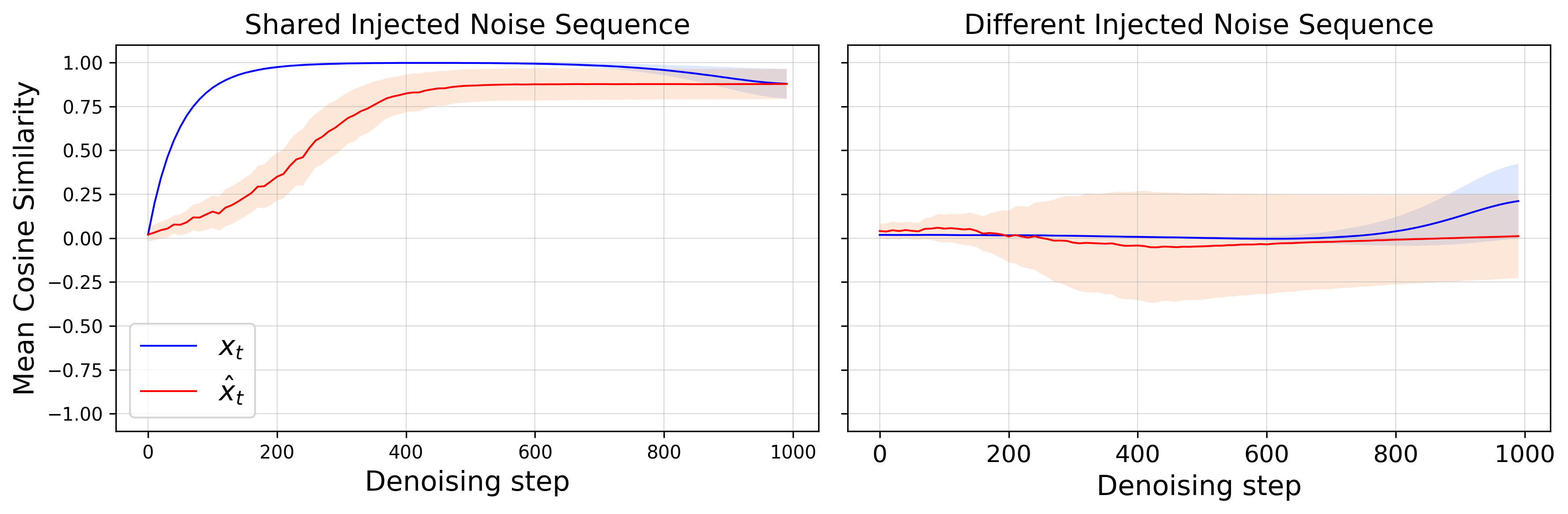}
    \caption{\textbf{Left}: Similarity rises during sampling; however, in this setup, it takes longer for $\mathbf{\hat{x}}_t$, as the model is likely less capable. \textbf{Right}: When different noise is injected, no similarity arises. Even in early states, the model is not able to pull in similar general directions for $\mathbf{\hat{x}}_t$.}
    \label{fig:butterflies_intermeidate_similarity}
\end{figure}

\begin{figure}[htbp]
    \centering
    \includegraphics[width=\linewidth]{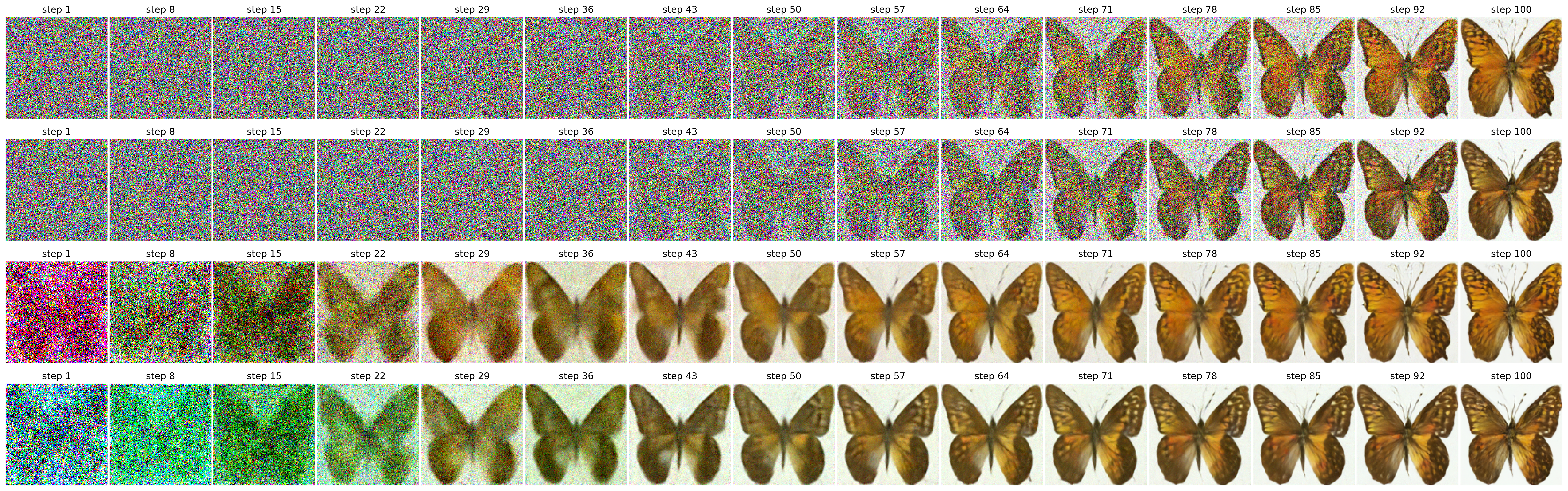}
    \caption{Sampling trajectories for model A and model B are shown. The top two rows show $\mathbf{x}_t$, and the bottom two rows show $\mathbf{\hat{x}}_t$, for the same injected noise with different starting noise conditions between model A and model B.}
    \label{fig:butterflies_trajectory}
\end{figure}

Measuring the similarity between intermediate states $\mathbf{x}_t$ and the states indicating the direction the model targets $\mathbf{\hat{x}}_t$ previously showed that similarity rises early during sampling when the same noise is injected (see Sec.~\ref{sec:similarity_during_sampling}). We performed the same experiment using our own models. While the overall trend of the results is similar, some notable exceptions are apparent (see Fig.~\ref{fig:butterflies_intermeidate_similarity} Left). The first is that similarity for $\mathbf{\hat{x}}_t$ rises significantly more slowly, indicating that the models struggle to identify the direction it needs to move in during the early stages of the process. Nevertheless, similarity rises to significant levels quickly for the sampling state $\mathbf{x}_t$ and then remains high until sampling finishes. The second difference is that no initial similarity for $\mathbf{\hat{x}}_t$ can be observed when different noise is injected (see Fig.~\ref{fig:butterflies_intermeidate_similarity} Right). This similarly suggests that the model yet struggled to move in a consistent direction with the large amounts of noise early on. This becomes even more apparent when visually inspecting the trajectories for $\mathbf{x}_t$ and $\mathbf{\hat{x}}_t$ (see Fig.~\ref{fig:butterflies_trajectory} Rows 3 and 4). During the early stages, even the supposedly denoised $\mathbf{\hat{x}}$ samples still contain high amounts of noise and only limited data. This is likely a limitation of the models due to the small training set.

\subsubsection{Regimes during sampling}
\begin{figure}[htbp]
    \centering
    \includegraphics[width=\linewidth]{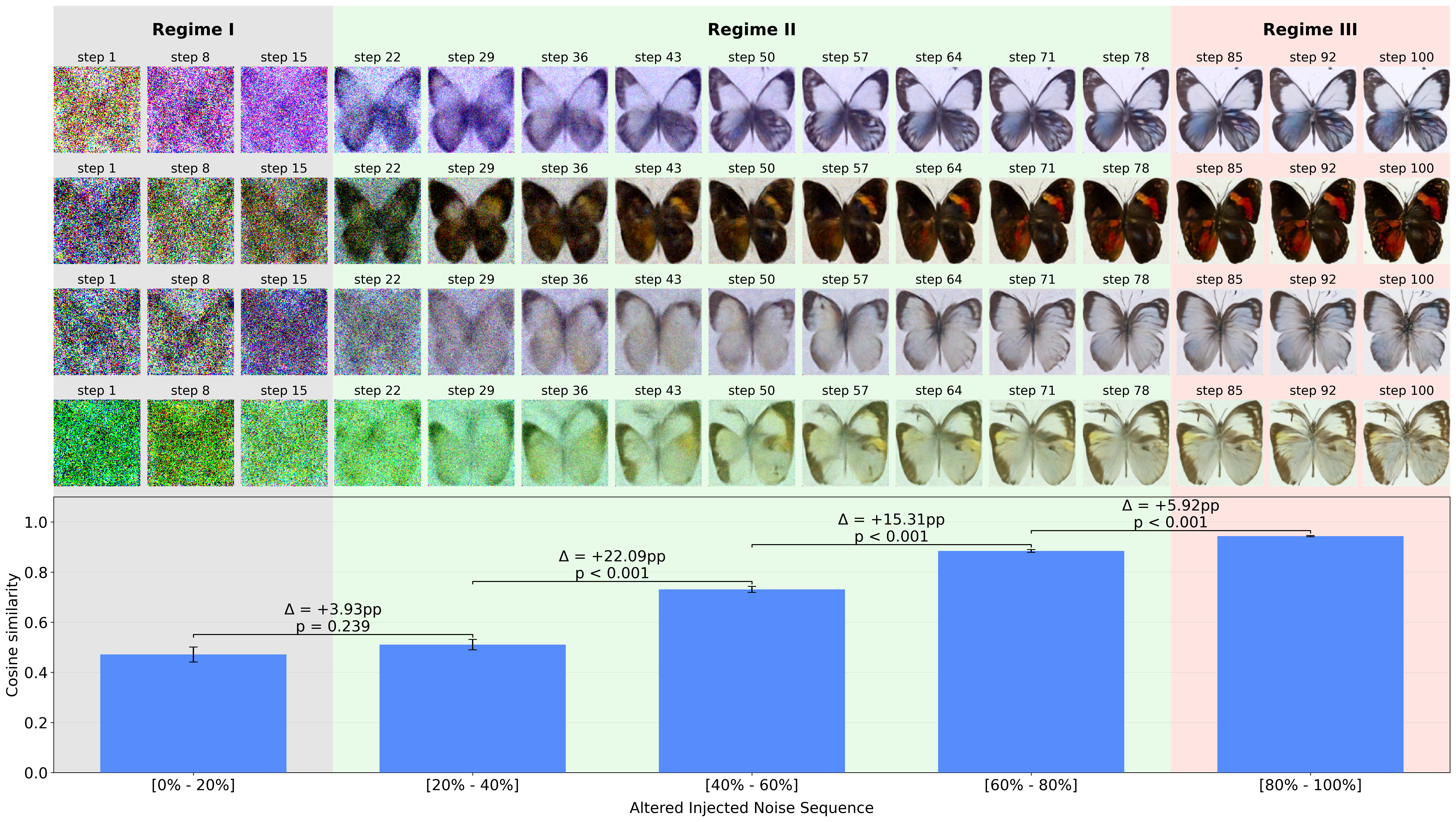}
    \caption{\textbf{Top}: Exemplary DM sampling trajectories (rows) using different injected noise sequences for the butterfly models. \textbf{Bottom}: The sensitivity to changes in the injected noise sequence rises throughout the process, early states do not show increased robustness, likely limited by the models capabilities ($p$-values show two-sided paired $t$-tests for adjacent intervals, $\alpha = 0.05$).}
    \label{fig:butterflies_regimes}
\end{figure}

In  previous experiments it already became clear that the models trained on the small set of training images are not able to denoise early states effectively (see Fig.~\ref{fig:butterflies_regimes} Regime I). Images in these early states still contain high amounts of noise and differ substantially between different runs, thus no increased robustness to changes could be observed when altering the first 20\% of injected noise of the sampling process. Robustness to changes in the sequence of injected noise then rises throughout the intervals (see Fig.~\ref{fig:butterflies_regimes}) similar to the previous results (see Fig.~\ref{fig:experiment_3_regimes_trajectories}) but plateauing earlier. Inspecting the samples visually also suggests that the process converged on a single sample over time, instead of switching targets in between as we observed in Section~\ref{fig:experiment_3_regimes_trajectories}.

\subsubsection{Ablation conclusion}
We could reproduce the first experiment in a different setting, showing that the injected noise plays an important role here as well. We could not identify the dynamical regimes as clearly and also found different behavior throughout the sampling process, which we attribute to the comparatively small dataset and the different model used compared to the main experiments. While successfully generating the final sample, this model requires many more, smaller steps that do not “wander around” as the model with more training does.

\subsection{Effects of initializing starting noise with a dataset sample}
Previous experiments showed that the injected noise is an important factor for the identity of the final sample when starting from pure noise. In this experiment, we investigated the effect of injected noise in cases with a higher signal-to-noise ratio (SNR). Given that the final phase of the diffusion process converges on a sample while the SNR reaches zero, we expected this experiment to give us insight into where in the process this phase starts. We used the same models as for the main experiments, trained on the CelebA dataset.

We initialized the sampling process with a vector that interpolates between Gaussian noise $\mathbf{z}_{\text{start}} \sim \mathcal{N}(\mathbf{0}, \mathbf{1})$ and a clean image from the dataset $\mathbf{x^*}$, $\mathbf{x}_0 = \alpha \mathbf{x^*} + (1 - \alpha) \mathbf{z}_{\text{start}}, \quad \alpha \in [0, 1]$. The injected noise was fixed across all runs. We then let the diffusion process run its course, in this case using the original SNR stopping criterion to avoid over-denoising the later samples with higher $\alpha$. We measured the cosine similarity to the sample that would be generated from pure noise ($\alpha = 0$) and to $\mathbf{x^*}$. The experiment was repeated for 100 trials.

\begin{figure}
  \centering
  \includegraphics[width=\linewidth]{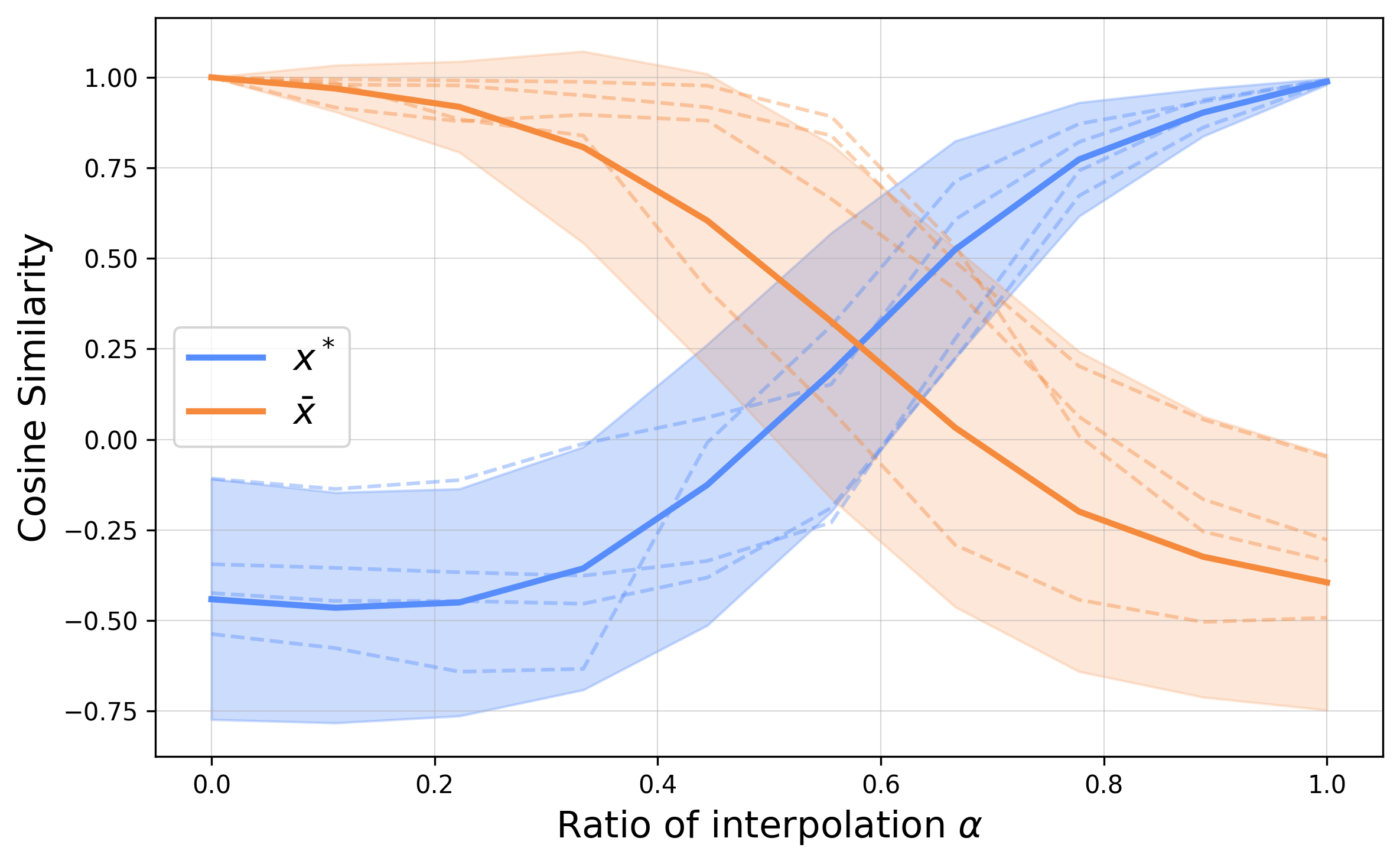}
  \caption{We initialize denoising with an increasing ratio between a clean image from the dataset and noise. Depicted are the cosine similarities to the clean image $\mathbf{x^*}$ and the image that would be generated from $\mathbf{z}_{\text{start}}$, with fixed injected noise. The dashed lines show exemplary similarities of individual samples (where the clean image is the same across different $\alpha$).}
  \label{fig:interpolating_start}
\end{figure}

The results (see Fig.~\ref{fig:interpolating_start}) show a sigmoidal tradeoff of similarity between the two final images, with a high variance. This variation may be dependent on the specific combination of injected noise and data sample $\mathbf{x^*}$.
Functionally, $\alpha$ is related to the time during the diffusion process $t$ in the sense that both correspond to a certain SNR (although the correspondence is nonlinear for $t$). Keeping this in mind, the results suggest that we can observe the transition between regime II and III here (as in regime III, the process would converge on $\mathbf{x^*}$ again), but the exact time of the transition is individual per sample.

\begin{figure}
  \centering
  \includegraphics[width=\linewidth]{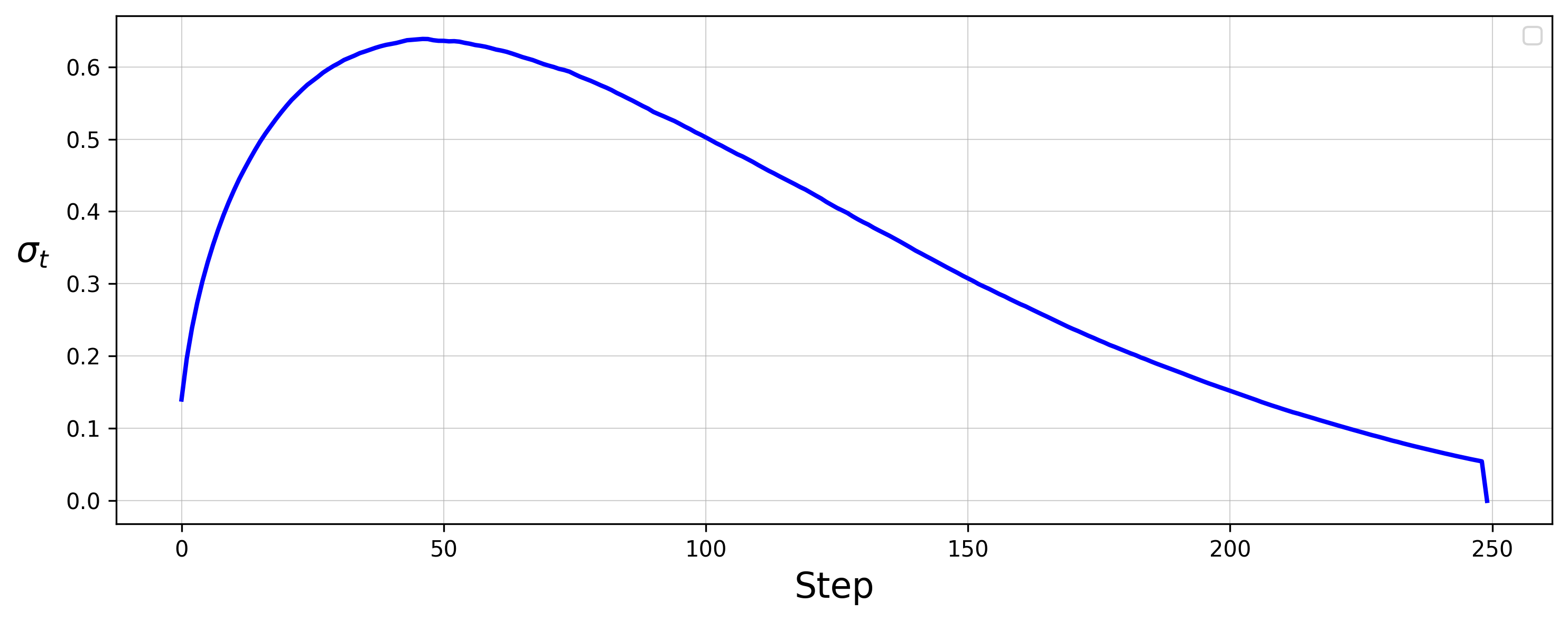}
  \caption{The average diffusion coefficient $\sigma_t$ during sampling, computed dynamically according to \cite{kadkhodaie2020solving}. This coefficient regulates the amount of injected noise based on the Signal-to-Noise-Ratio of the current state.}
  \label{fig:noise_schedule}
\end{figure}

\end{document}